\documentclass[journal]{IEEEtran}
\usepackage{amsmath,amsfonts}
\usepackage{algorithm}
\usepackage{array}
\usepackage[caption=false,font=normalsize,labelfont=sf,textfont=sf]{subfig}
\usepackage{textcomp}
\usepackage{stfloats}
\usepackage{url}
\usepackage{verbatim}
\usepackage{graphicx}
\usepackage{cite}
\usepackage{amssymb}
\usepackage{booktabs}
\usepackage{algorithmicx}
\usepackage{algpseudocode} 
\usepackage{bm}
\usepackage{multirow}
\usepackage{nameref}
\usepackage{float}
\usepackage{makecell}
\usepackage{xcolor}
\usepackage[capitalize]{cleveref}

\begin{document}

\title{Ground Plane Matters: Picking Up Ground Plane Prior in Monocular 3D Object Detection}
\author{Fan Yang, Xinhao Xu, Hui Chen, Yuchen Guo, Jungong Han,~\IEEEmembership{Senior Member,~IEEE}, \\ Kai Ni, and Guiguang Ding,~\IEEEmembership{Member,~IEEE}
\thanks{Fan Yang, Xinhao Xu, Hui Chen, Yuchen Guo and Guiguang Ding are with the Beijing National Research Center for Information
Science and Technology, Tsinghua University, Beijing,
100084, China.
Fan Yang, Xinhao Xu and Guiguang Ding are also with the School of Software, Tsinghua University, Beijing,
100084, China.
Jungong Han is with the Department of Computer Science, Aberystwyth University, Aberystwyth, SY23 2AX, UK.
Kai Ni is with the HoloMatic Technology, Beijing, 100102, China.}
\thanks{Fan Yang and Xinhao Xu have equal contribution. \emph{(Corresponding author:
Guiguang Ding.)}
(e-mail: yfthu@outlook.com; dinggg@tsinghua.edu.cn)}
}

\maketitle

\begin{abstract}
   The ground plane prior is a very informative geometry clue in monocular 3D object detection (M3OD). However, it has been neglected by most mainstream methods. In this paper, we identify two key factors that limit the applicability of ground plane prior: the projection point localization issue and the ground plane tilt issue. To pick up the ground plane prior for M3OD, we propose a Ground Plane Enhanced Network (GPENet) which resolves both issues at one go. For the projection point localization issue, instead of using the bottom vertices or bottom center of the 3D bounding box (BBox), we leverage the object's ground contact points, which are explicit pixels in the image and easy for the neural network to detect. For the ground plane tilt problem, our GPENet estimates the horizon line in the image and derives a novel mathematical expression to accurately estimate the ground plane equation. An unsupervised vertical edge mining algorithm is also proposed to address the occlusion of the horizon line. Furthermore, we design a novel 3D bounding box deduction method based on a dynamic back projection algorithm, which could take advantage of the accurate contact points and the ground plane equation. Additionally, using only M3OD labels, contact point and horizon line pseudo labels can be easily generated with NO extra data collection and label annotation cost. Extensive experiments on the popular KITTI benchmark show that our GPENet can outperform other methods and achieve state-of-the-art performance, well demonstrating the effectiveness and the superiority of the proposed approach. Moreover, our GPENet works better than other methods in cross-dataset evaluation on the nuScenes dataset. Our code and models will be published.
\end{abstract}

\begin{IEEEkeywords}
3D object detection, monocular images, autonomous driving.
\end{IEEEkeywords}

\section{Introduction}
\IEEEPARstart{P}{erception} of 3D spatial information is critical for autonomous driving systems. Many methods rely on LiDAR to implement 3D object detection. However, LiDAR sensors are not cost-efficient for resource-limited commercial autonomous driving systems due to their high expense. Thanks to the breakthrough in computer vision, directly inferring the spatial information of objects from images becomes appealing and has obtained remarkable progress recently\cite{bao2019monofenet,liu2022fine,xu2021multi,ye2021unsupervised,zhou2019unsupervised,song2021mlda}. Researchers have developed monocular 3D object detection (M3OD) technique, which has the potential to approach LiDAR's performance with a cheap camera \cite{mousavian20173d,monogrnet,smoke,gupnet}.

Although it has yielded promising accuracy, M3OD is still confronted with a crucial challenge, i.e. inferring 3D depth information from a single 2D input.
In fact, how to estimate the missing dimension of depth from the input 2D image has long been an intractable problem for the computer vision community. Considerable efforts have been put into finding solutions that introduce geometry constraints into M3OD to tackle this problem.

One simple strategy is to estimate the 2D and 3D height of a specific object, denoted as $h_{2D}$ and $h_{3D}$, respectively. Then the depth can be simply computed by $d=f_y \cdot h_{3D}/h_{2D}$ , where $f_y$ is the focal length of the camera in Y-axis \cite{wang2021depth,ku2019monocular}.
However, it is difficult to acquire $h_{3d}$ precisely in real scenarios where the 3D size of objects varies a lot. Besides, $h_{2d}$ used in the depth estimation is not equal to the height of the 2D BBox, resulting in great difficulty to infer the exact $h_{2d}$ from images \cite{gupnet,BarabanauABM20}.
Another strategy detects the keypoints of the object (usually the vertices of the 3D BBox) and derives the object's location through the keypoints projection \cite{chabot2017deep, li2020rtm3d}. 
But the projected 3D BBox vertices in the 2D image have no clear semantics, leading to difficulties in precisely localizing such vertices. Also, purely relying on the keypoints inside one object will lose the geometric clues corresponding to other objects or environments.
\begin{figure}
	\begin{center}
		\includegraphics[width=200pt]{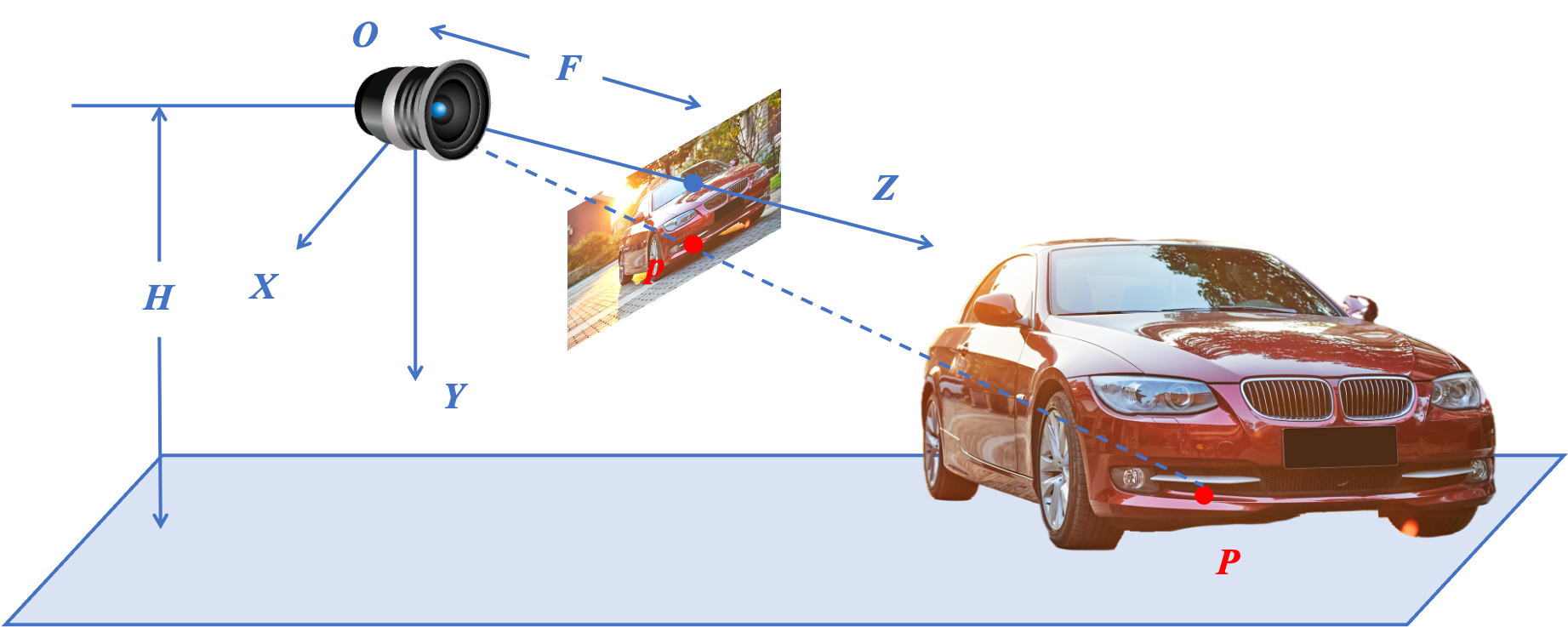}
	\end{center}
	\vspace{-0.2cm}
	\caption{\textbf{Illustration of the back projection with the ground plane.}
	$H$ is the camera's height and $F$ is the camera's focal length. $p$ is a pixel in the image and $P$ is the spatial position after back projection. The blue plane in the figure represents the ground plane.}
	\label{fig:intro1}
	\vspace{-0.5cm}
\end{figure}
Moreover, there is a geometric strategy introduced in  VisionACC~\cite{stein2003vision}, which adopts ground plane prior and the back projection
to accomplish the 3D localization.
\IEEEpubidadjcol
Specifically, as shown in Fig. \ref{fig:intro1}, each pixel in the image corresponds to a ray in space. For any pixel, which is the image projection of a point on the ground, its spatial location can be obtained by finding the intersection of the ray and the ground plane. 
The corresponding relation between the ground plane and the object can provide a lot of geometric and spatial clues. 

{
\begin{figure}
\vspace{-0.1cm}
	\begin{center}
		\includegraphics[width=\linewidth]{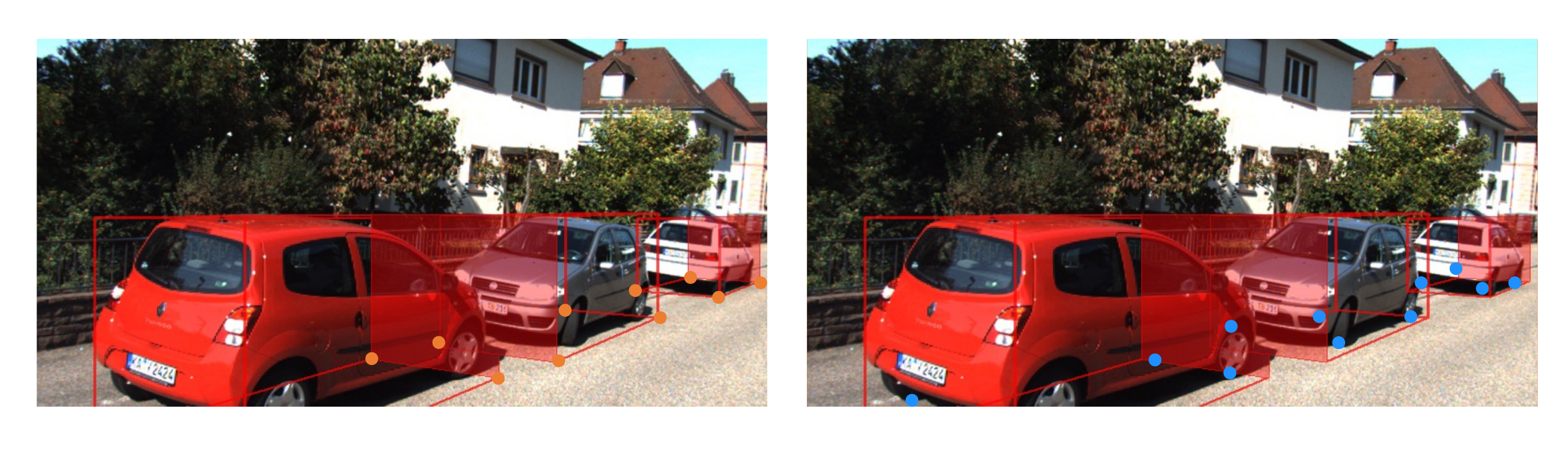}
	\end{center}
		\vspace{-0.3cm}
	\caption{\textbf{Comparison between the 3D BBox's bottom vertices (orange dots) and ground contacting points (blue dots).} Obviously, the ground contact points have stronger semantic information than the bottom vertices.}
		\vspace{-0cm}
	\label{fig:intro3}
		\vspace{-0cm}
\end{figure}
}

\begin{figure}
	\begin{center}
		\includegraphics[width=247pt]{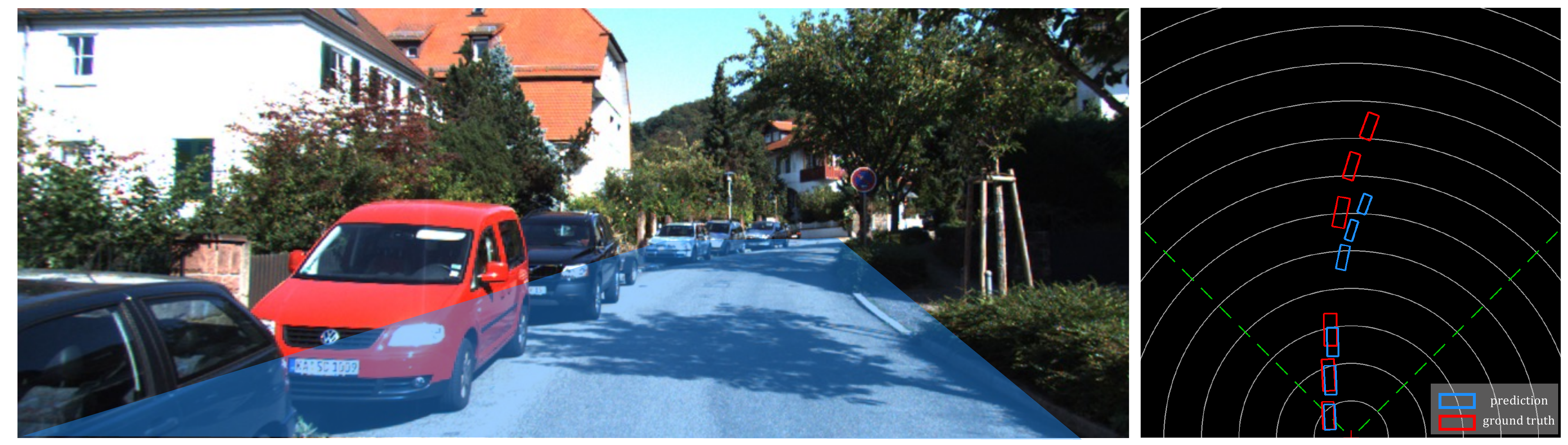}
	\end{center}
		\vspace{-0.2cm}
	\caption{\textbf{The estimation deviation caused by the ground plane tilt problem.} The blue plane is the preset fixed ground plane, neglecting the ground plane's tilt. From the bird's eye view (BEV), the inferred 3D BBox (blue box) using the fixed ground plane will gradually drift from the ground truth (red box).}
	\label{fig:intro2}
	\vspace{-0.2cm}
\end{figure}

Notwithstanding, the ground plane prior has been generally neglected by mainstream M3OD methods in this era. 
We argue that the following two major drawbacks can account for the recession in the ground plane projection method. 
(1) The projection point localization issue. In the critical ground projection process, existing works \cite{groundaware,gupta20183d} use the 3D BBox bottom vertices or the bottom center point as the projection reference point. However, as illustrated in Fig. \ref{fig:intro3}, such points do not have clear semantics in images and they are often too rough to be learned precisely by neural networks. The coarse point may result in fluctuations in the estimation of the object's spatial position.
(2) The ground plane tilt problem. In autonomous driving scenes, the ground plane in the camera coordinate system (CCS) is not static and the pitch and roll of the ground may change owing to the ego car bumping. It can change dramatically when the ego car or the road fluctuates. However, previous works \cite{groundaware, qin2022monoground} usually employ a fixed ground plane and neglect the ground plane's tilt in the CCS.
As a result, large errors would occur when using an inaccurate ground plane reference to infer objects' spatial position (Fig. \ref{fig:intro2}).

\begin{figure*}[htbp]
	\begin{center}
		\includegraphics[width=\linewidth]{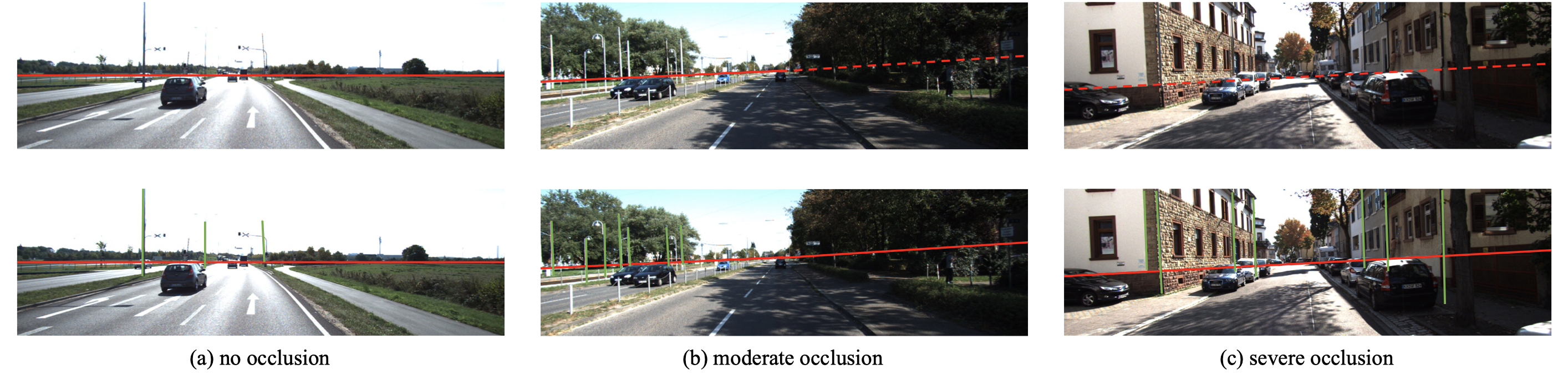}
	\end{center}
	\vspace{-0.2cm}
	\caption{\textbf{Horizon line detection.} Our vertical edge detection (in green) can be applied as an assistance to the horizon line detection.}
	\vspace{-0.3cm}
	\label{fig:horizon}
\end{figure*}

To tackle the issues above, we propose a \textbf{G}round \textbf{P}lane \textbf{E}nhanced Network, namely GPENet, which picks up the ground plane constraint to improve the performance of M3OD. Firstly, for the projection point localization issue, compared to the implicit 3D BBox bottom vertices or center, we observe that the contact points between objects and the ground plane are explicit pixels and direct information in images (Fig. \ref{fig:intro3}). They are rich in semantics, easy for neural networks to detect precisely, and could be directly projected with the ground plane.
Secondly, to address the problem of ground plane tilt, the idea is to infer the ground plane equation for each image, and then perform an accurate back projection from 2D to 3D space. Note that directly inferring the ground plane equation from the 2D image is considerably difficult for the neural network. We find that in principle, to capture such dynamic ground plane, we can leverage a point, i.e., the camera's optical center, and a line, i.e., the horizon line~\footnote{The horizon line is defined as the projection of the ground plane in the image when its depth approaches infinity.}, to deduce such plane mathematically. Intuitively, compared to the ground plane, the horizon line is easier to deduce for the neural network.
However, the problem arises in practical situations, where the horizon is usually occluded (Fig. \ref{fig:horizon} c). In this case, the neural network could not work properly to detect such line. To remedy this situation, we further design a novel vertical edge mining algorithm which can enhance the horizon line detection with the salient vertical edges in the image like buildings' edge, cement column, lamp pole and so on(Fig. \ref{fig:horizon}). Eventually, with the precise image horizon line enhanced by image vertical edges, we derive the accurate and dynamic ground plane equation mathematically and the depth deviation caused by the ground plane tilt problem could be adequately alleviated.

Based on the detected contact points of the object and the deduced dynamic ground plane equation, we propose a dynamic back projection algorithm which could derive the object's various 3D attributes, including position, dimension and rotation. Our 3D BBox deduction scheme would achieve high precision M3OD.
Furthermore, thanks to our exquisite pseudo labels generation method, labels used for training ground contact points and horizon lines can be generated only using M3OD labels, with no need for extra data and no labelling cost.

In summary, our main contributions are as follows.

(1) We identify two key factors hindering the widespread application of traditional ground plane prior: the projection point localization issue and the ground plane tilt issue. Therefore, we propose a Ground Plane Enhanced Network, namely GPENet, to pick up the significant ground plane geometry constraint for M3OD.

(2) For the projection point localization issue, we propose to leverage the ground contact points between the objects and the ground, which are explicit and direct pixels in images. Such points are more suitable for neural networks to localize precisely. 

(3) To tackle the ground plane tilt issue, we propose to estimate the ground plane equation based on the horizon line in the image. To deal with the occlusion problem of the  horizon line in real scenes, we further design a novel algorithm which extracts vertical edges in images to enhance the horizon line detection, accomplishing a robust ground plane estimation. 

(4) We put forward a dynamic back projection algorithm, which employs the proposed precise contact points and dynamic ground plane equation. Besides the depth, various 3D attributes also could be determined by geometry deduction, including objects' 3D size and rotation. An elaborate 3D bounding box generation scheme is further introduced based on dynamic back projection.

(5) On the most commonly used benchmark KITTI, our method achieves the state-of-the-art performance without using extra data. The experimental results on nuScenes dataset show the great generalization capacity of our method. Extensive experiments demonstrate the effectiveness and emphasize the importance of ground plane prior, which could be a very informative and potential geometry clue for M3OD.
\section{Related Work}

\subsection{Keypoints in Monocular 3D Object Detection}
Monocular 3D object detection aims to infer the 3D BBox of the objects from a single RGB image \cite{ding2020learning,he2019mono3d}.
To obtain the accurate 3D bounding box, keypoints have been employed as an auxiliary information in M3OD, which can also assist in inferring the shape of occluded and truncated objects.
F. Chabot et al.~\cite{chabot2017deep} use a 3D CAD model with a fixed number of keypoints. They predict the CAD template similarity and match the model with the highest similarity to reconstruct the object's shape and orientation. 
CAD models have shown an effect in the shape detection. But other 3D attributes' results, especially the depth prediction, are still far from satisfactory. 

Some works use keypoints to conduct projections. For example, RTM3D~\cite{li2020rtm3d} uses a feature pyramid network (FPN)~\cite{lin2017feature} as the feature extraction backbone, and a multi-task detection head to predict different attributes of objects. RTM3D uses nonlinear least square optimization to minimize the distance between the 3D position projection and the 2D keypoints predicted by the neural network.
KM3D~\cite{li2021monocular} improves RTM3D by embedding the 2D-3D geometric constraints into the training process of the neural network, and matrix calculus is used for gradient back propagation to minimize the re-projection loss. 
MonoFlex \cite{Zhang_2021_CVPR} uses keypoints to assist the process of depth estimation and adopts an uncertainty guided ensemble method to improve its accuracy.
These works use the vertices of a 3D BBox to establish the geometric projection, which is hard for neural networks to detect precisely due to the lack of direct semantics.

\subsection{Geometry Projection in Monocular 3D Object Detection} 
Recently, many works attempt to involve geometric priors in M3OD, achieving encouraging results \cite{monogrnet,ma2020rethinking}.
Deep3DBox~\cite{mousavian20173d} uses neural networks to predict an object's orientation, dimension and 2D bounding box to provide constraints for the 3D bounding box. 
Some works \cite{groundaware,qin2022monoground} use the ground plane to assist M3OD, but they overlook the ground plane tilt problem and use the rough 3D BBox bottom center to conduct projections. In addition, MonoRCNN \cite{shi2021geometry} introduces the geometric information between 2D BBox height and 3D height to estimate the objects' depth. 
GUPNet~\cite{gupnet} also estimates objects' depth with objects' 2D and 3D height projection and uses uncertainty loss to determine objects' score more precisely. MonoEF\cite{zhou2021monocular} focuses on the changes of camera external parameters, and uses the extra data, i.e., the odometry datasets to train the network. It designs a feature transfer network to rectify the feature disorder. Unlike these implicit feature transfer networks, our method finds a way to calculate the ground plane equation directly.

\section{Approach}
\begin{figure*}[htbp]
	\begin{center}
		\includegraphics[width=\linewidth]{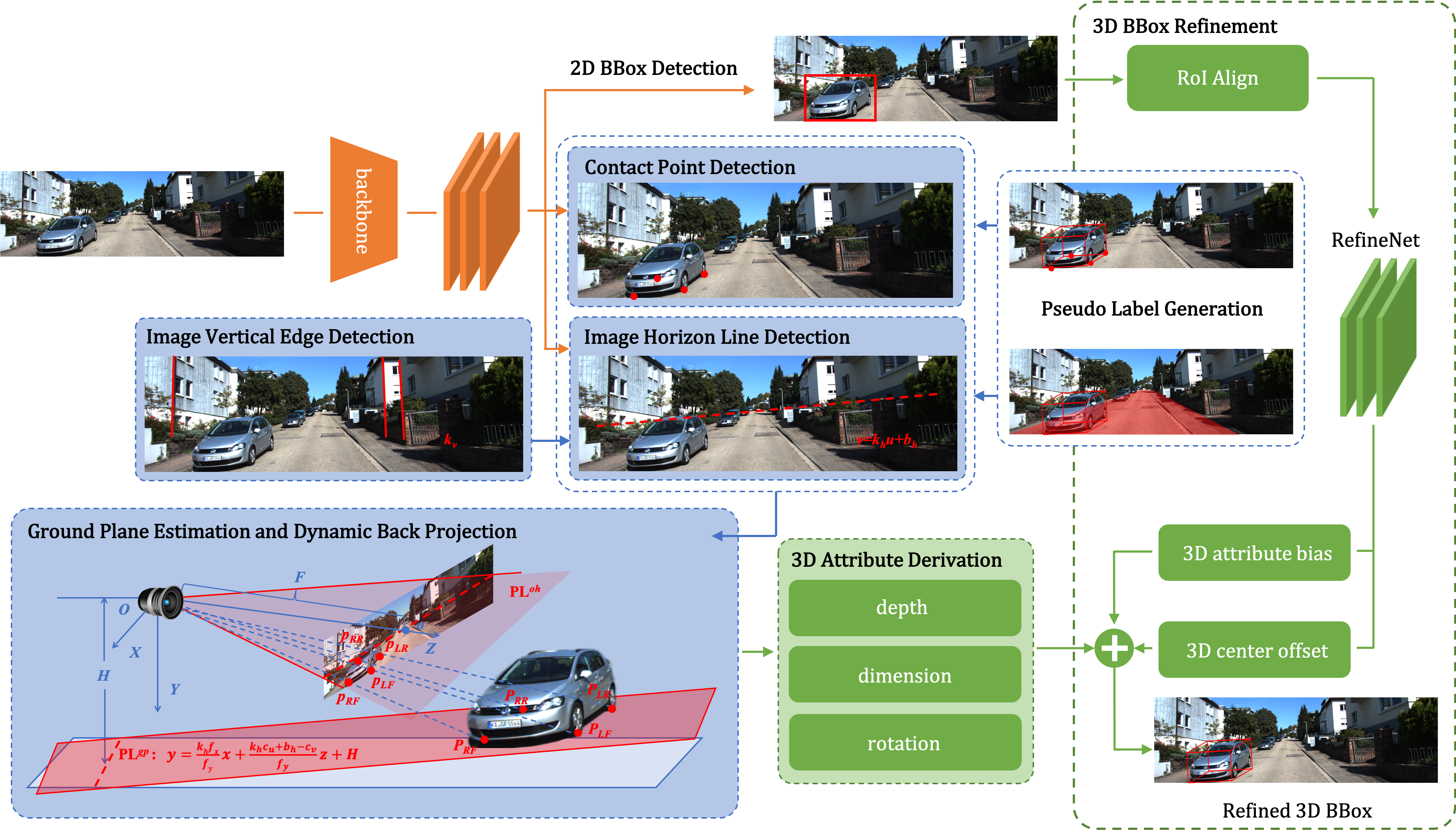}
	\end{center}
		\vspace{-0.2cm}
	\caption{\textbf{The framework of the Ground Plane Enhanced Network.} We use CenterNet \cite{centernet} to predict the 2D BBox, ground contact points and the horizon line from the input image. Next, the ground plane equation ${\rm PL}^{gp}$ (Eq. \ref{P2}) is derived from the detected image horizon line mathematically. Then with the contact points and the ground plane equation, dynamic back projection is conducted to infer a geometry-based 3D BBox. Finally, the RefineNet is employed to refine the 3D BBox. Additionally, we design the pseudo label generation scheme using only M3OD labels to train the contact point and the horizon line detection head, without any extra data.}
	\label{fig:pipeline}
		\vspace{-0.3cm}
\end{figure*}
\subsection{Overview}

The key point of our method is to pick up the ground plane prior to improve the monocular 3D object detection. The conventional procedure consists of two parts: (1) 2D object localization which aims to localize the object and projection point in the image. (2) Back projection guided by the ground plane. Although good progress has been made, this procedure faces two major difficulties. Firstly, in 2D localization part, existing methods use 3D BBox vertices or bottom center as projection points which are rough and have no clear semantics in the image. Therefore, it is very difficult to precisely perceive such points for the neural network. Moreover, previous works \cite{groundaware} simply employ a fixed ground plane as the projection plane, which takes no consideration of the ground plane tilt problem, resulting in large deviation. For the first issue, we propose a contact points guided 2D object localization method (Section \ref{sec:2D Object Localization Guided by Contact Points}) to realize the accurate positioning of projection points. In order to solve the ground plane tilt problem, we design an ingenious vertical-edge-enhanced horizon line detection algorithm and propose to estimate the exact ground plane equation mathematically based on the horizon line (Section \ref{sec:Ground Plane Estimation Based on Horizon Line}). Finally, we design a 3D attribute deduction method based on dynamic back projection, which can achieve accurate 2D-3D inference (Section \ref{sec:3D Bounding Box Deduction Based on Dynamic Back Projection}).

\begin{figure}
	\begin{center}
		\includegraphics[width=\linewidth]{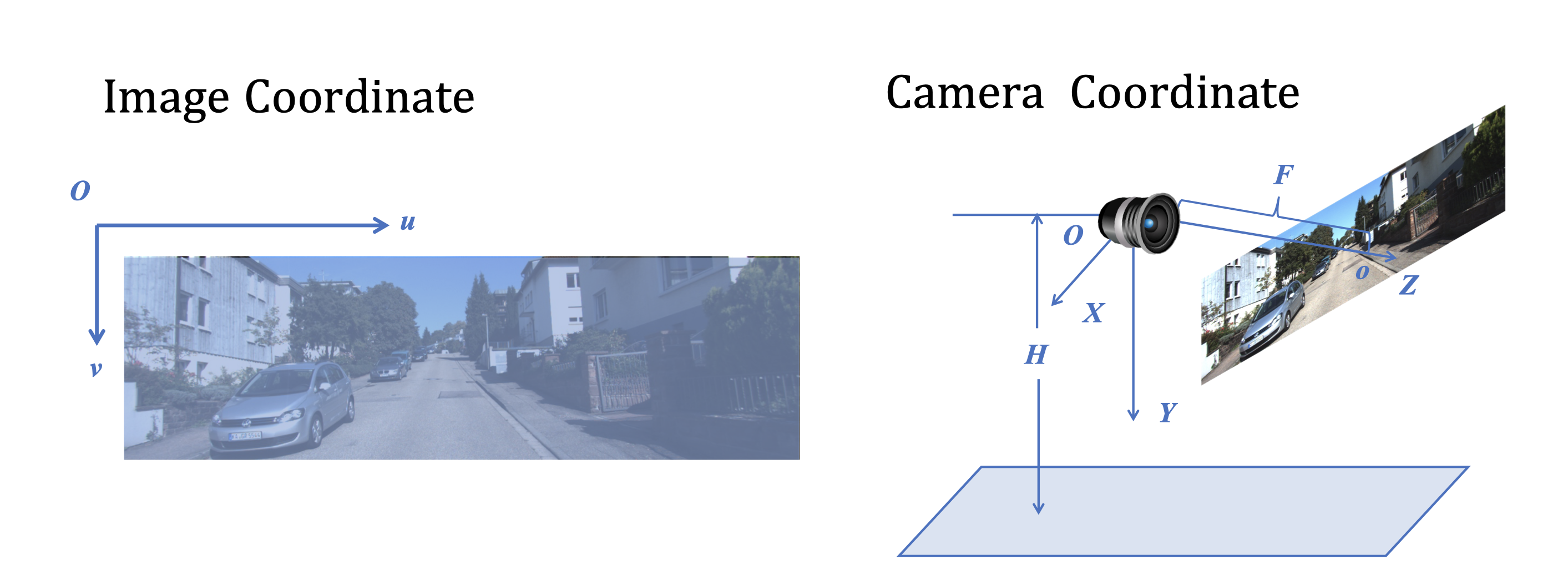}
	\end{center}
		\vspace{-0.25cm}
	\caption{\textbf{The image coordinate and the camera coordinate.} The left subfigure is the image coordinate: the origin point is in the upper left corner of the image with u-axis right and v-axis down. The right subfigure is the camera coordinate system (CCS): the origin point is the camera's optical center with X-axis right, Y-axis down and Z-axis forward along the optical axis of the camera.}
	\vspace{-0.4cm}
	\label{fig:coord}
\end{figure}

Note that the horizon refers to the horizon line in the image coordinate by default, and the ground plane refers to the ground plane in 3D coordinate by default. The image coordinate system, the camera coordinate system (CCS) and their axis' direction are shown in Fig. \ref{fig:coord}.

\subsection{2D Object Localization Guided by Contact Points}
\label{sec:2D Object Localization Guided by Contact Points}
\subsubsection{Contact Point Detection}
\label{sec:Contact Points Detection}
We begin with the localization of objects in the 2D image. Rather than the 3D BBox vertices or bottom center used by most previous works \cite{mousavian20173d, li2020rtm3d, li2021monocular}, we propose to use the ground contact points as the projection points, because they have clear semantics in the image (Fig. \ref{fig:intro3}) and are much easier to be detected by neural networks. 
In order to achieve the ground contact point detection, we refer to the keypoints detection in CenterNet \cite{centernet} and detect $k$ keypoints for each object instance. $k$ represents the number of one object's contact points, i.e., $k=4$ for cars, $k=2$ for pedestrians and cyclists. The contact point detection is comprised of three detection heads. The first head's output $\widehat{Y}$ is the heatmap of all objects' contact points in the image. The second head outputs the contact points' local offset $\widehat{Y}_{of}$ to remedy the down sampling error and the third head's output $\widehat{Y}_c$ represents the indication from the object center.
Defining $Y, Y_{of}, Y_c$ as their ground truths, the keypoints loss is as follows:
\begin{equation}
\begin{split}
L_{kps} = L_{kps,hm} + L_{kps,of} + L_{kps,C} \quad \quad \quad \quad \quad \ \ \\
 = {\rm Focal} (\widehat{Y}, Y) + {\rm L_1} (\widehat{Y}_{of}, Y_{of}) + {\rm L_1} (\widehat{Y}_c, Y_c) \ 
\end{split}
\end{equation}
where $L_{kps,hm}$ is the focal loss \cite{focalloss} of the keypoint heatmaps, $L_{kps,of}$ is the $\rm L_1$ loss of the points' local offset and $L_{kps,C}$ is the $\rm L_1$ loss of the indication from the object center.

	\vspace{-0cm}
The focal loss \cite{focalloss} is as follows:
\begin{equation} 
\begin{split}
&{\rm Focal} (\widehat{T}, T) \\
& = \frac{-1}{UVC} \sum_{uvc}\begin{cases}
(1-\widehat{T}_{uvc})^\alpha \log (\widehat{T}_{uvc}) & {\rm if} \ T_{uvc} = 1 \\
(1-T_{uvc})^\beta \widehat{T}_{uvc}^\alpha \log (1-\widehat{T}_{uvc}) & {\rm otherwise}
\end{cases}
\end{split}
\label{focalorigin}
\end{equation}
where $\widehat{T}$ is the predicted tensor and $T$ is the ground truth. $U,V,C$ are the tensor's size and $u, v, c$ are the indexes. We set focal loss parameters $\alpha = 2, \beta = 4$. The $\rm L_1$ loss is as follows:
\begin{equation}
{\rm L_1} (\widehat{T}, T) = {\rm mean}(\left| \widehat{T} - T \right|)
\end{equation}
where $\widehat{T}$ is the predicted value and $T$ is the ground truth.

Benefited from the well-designed keypoint detection heads, our method is capable of detecting the ground contact points accurately even when they are occluded. We use $(u_{CP},v_{CP})^\top$ to denote one of the detected ground contact points in the image, and we will back project $(u_{CP},v_{CP})^\top$ to the CCS in Section \ref{sec:Dynamic Back Projection of Contact Points With Ground Plane Equation}.
\subsubsection{Contact Point Pseudo Label Generation}
\label{sec:Contact Point Pseudo Label Generation}
\begin{figure}
	\begin{center}
		\includegraphics[width=\linewidth]{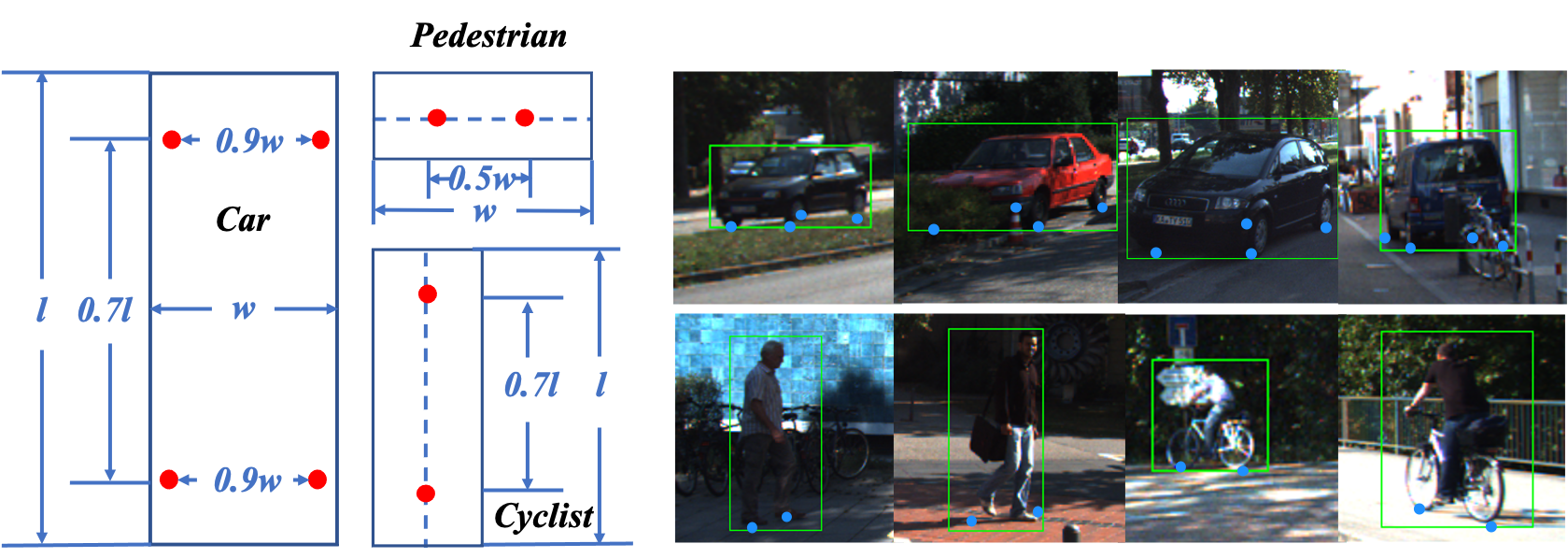}
	\end{center}
	\vspace{-0.2cm}
	\caption{\textbf{Ground contact point pseudo label generation.} We design the pseudo label generation scheme using 3D BBox labels. Specifically, the position of the ground contact points satisfies a proportional relationship with the size of the 3D BBox. The left figure shows the detailed relationship in BEV and the right one is the visualization of pseudo labels.
	}
	\vspace{-0.3cm}
	\label{fig:plabel}
\end{figure}

One non-negligible issue in contact point detection is that the M3OD dataset does not annotate the ground contact point labels. Thus, we propose a contact point pseudo label generation scheme to train our network. Specifically, we first localize the ground contact points guided by the object's 3D BBox in the CCS. We determine the spatial position of 3D BBox's bottom face and then utilize the wheelbase information to localize the contact points.
Taking Car category for example, we define $k_l$ as the ratio of the front and rear wheels' distance to the length of the 3D BBox and $k_w$ as the ratio of the left and right wheels' distance to the width of the 3D BBox. Both $k_l$ and $k_w$ can be obtained with the cars' average wheelbase ($k_l=0.7$, $k_w=0.9$ in KITTI, as shown in Fig. \ref{fig:plabel}). 
We define $P^o_{LF}, P^o_{RF}, P^o_{RR},$ $P^o_{LR}$ ($LF$, $RF$, $RR$ and $LR$ represent the left-front, right-front, right-rear and left-rear point, respectively) as the four contact points in the object's local coordinate system\footnote{We define that object local coordinate's axis direction is the same as the camera coordinate, i.e., X-axis right, Y-axis down, Z-axis forward  along the main optical axis of the camera. The origin point of the object local coordinate is the center of the bottom face and the front of the object faces the positive direction of the X-axis.}:
\begin{equation}
\begin{aligned}
P_{LF}^o =  [\frac{k_l}{2} l, 0, \frac{k_w}{2} w]^\top, \ P_{RR}^o = [- \frac{k_l}{2} l, 0,  -\frac{k_w}{2} w]^\top \\
P_{LR}^o =  [- \frac{k_l}{2} l, 0, \frac{k_w}{2} w]^\top, \ 
P_{RF}^o = [\frac{k_l}{2} l,  0, - \frac{k_w}{2} w]^\top
\end{aligned}
\end{equation}
where $l$ and $w$ are the length and width of the object, respectively.

Given one coordinate $P^o$ in the object's local coordinate system, the position $P^c$ in the CCS can be obtained by using the rotation matrix $\mathbf{R}$ and translation vector $\mathbf{T}$:
\begin{equation}
\begin{aligned}
& \quad \quad \quad \quad \quad \quad P^c =\mathbf{R} P^o + \mathbf{T} \\
& \mathbf{R} = \begin{bmatrix} cos\theta & 0 & - sin\theta \\ 0 & 1 & 0 \\ sin\theta & 0 & cos\theta \end{bmatrix}, \quad  
\mathbf{T} = \begin{bmatrix} x, y, z \end{bmatrix}^{T}
\end{aligned}
\end{equation}
where $x,y,z$ are the position of the object's bottom center in the camera coordinate and $\theta$ is the corresponding rotation angle around Y-axis. With the camera intrinsic matrix $\mathbf{K}$, we can derive the image pixel coordinates $p = [u, v]^\top$ of the contact point:
\begin{equation}
\begin{aligned}
& z_p [u, v, 1]^\top = \mathbf{K}P^c = \mathbf{K} (\mathbf{R} P^o + \mathbf{T}) \\
& \ \quad \quad \ \ \mathbf{K} = \begin{bmatrix} f_{x} & 0 & c_{u} \\ 0 & f_{y} & c_{v} \\ 0 & 0 & 1 \end{bmatrix} \label{K}
\end{aligned}
\end{equation}
where $z_p$ is the Z-axis coordinate of $P^c$ in the CCS and $f_{x},f_{y},c_{u}, c_{v}$ are camera intrinsic parameters. When $P^o$'s value is set to $P^o_{LF}, P^o_{RF}, P^o_{RR}$ or $P^o_{LR}$, the corresponding contact point's image pixel $p_{LF}, p_{RF}, p_{RR}$ or $p_{LR}$ can be derived.

For the other categories, we generate contact point pseudo label according to their geometric location of grounding points and Fig. \ref{fig:plabel} shows the implementation details and the effects of pseudo label generation.

\subsubsection{2D Bounding Box Detection}
In the 2D BBox detection part, we employ CenterNet \cite{centernet} and predict the object's center heatmap $\widehat{B}$, the center offset $\widehat{B}_{of}$ and the 2D BBox size $\widehat{B}_s$. We use focal loss \cite{focalloss} for the center heatmap, $\rm L_1$ loss for center offset and box size. The 2D BBox detection loss $L_{2d}$ is as follows:
\begin{equation}
\begin{split}
L_{2d} = L_{center} + L_{2d\_of} + L_{2d\_size} \quad \quad \quad \quad \quad \\
= {\rm Focal} (\widehat{B}, B) + {\rm L_1} (\widehat{B}_{of}, B) + {\rm L_1} (\widehat{B}_s, B) 
\end{split}
\end{equation}
where $B$, ${B_{of}}$, $B_s$ are the center heatmap, center offset and 2D BBox size ground truths, respectively.
\subsection{Ground Plane Estimation Based on the Horizon Line}
\label{sec:Ground Plane Estimation Based on Horizon Line}
\subsubsection{Analysis}
For the second issue hindering the use of ground plane, i.e., the ground plane tilt issue, we estimate a mathematical formulation of the ground plane to solve the problem.
Note that directly estimating such tilt from the image is too obscure for neural networks and our solution is to leverage the horizon line in the image. As shown in Fig. \ref{fig:horizon}, the horizon is the image projection of the ground plane at infinity, where the sky seems to meet the land. Such horizon information can reflect the ego pose of the camera relative to the ground. When the pitch angle of the camera changes, the horizon moves up and down in the image. When the roll angle of the camera changes, the horizon rotates in the image.
Therefore, we aim to leverage such horizon information to derive a accurate ground plane estimation method mathematically for M3OD.

Note that the naive image horizon detection dose not work in the environment where the horizon is not clear. As shown in Fig. \ref{fig:horizon} (c), the horizon line in the image is severely occluded owing to the obstacles in the scene, which makes the horizon implicit and difficult to detect. In order to estimate the ground plane in the image with many obstacles, 
we draw inspiration from human's recognition pattern. The pattern is that humans usually utilize large objects in the scene like buildings, cement columns and lamp poles as guidance to infer ego posture and the tilt of the ground plane. We just imitate such human's capacity and go deep into the details. Intuitively, the salient edges in images, especially vertical edges, can reflect the rotation of the scenes. Specifically, the man-made buildings tend to be upright so most vertical edges of these buildings are orthogonal to the horizon line. Therefore, we design an ingenious unsupervised digital image processing algorithm to extract salient vertical edges in images and dig out the valuable edges. Then we deduce the slope of the vertical edges and the horizon line. Finally, we can complete the horizon line detection and ground plane estimation whether it is occluded or not.
\subsubsection{Image Horizon Line Detection}
In order to accomplish the 2D horizon line detection task, we add a heatmap head in the 2D detection part of the network. 
It is a pixel-level heatmap between 0 and 1, indicating the probability of a pixel in the horizon line.
We adopt the focal loss~\cite{focalloss} during the process of training. 
\begin{equation}
\begin{split}
L_{hor} = {\rm Focal}(\widehat{M}, M)
\end{split}
\label{Lhor}
\end{equation}
where $\widehat{M}$ is the predicted horizon heatmap and $M$ is the ground truth. Note that Eq. \ref{focalorigin} shows the details of focal loss and in Eq. \ref{Lhor}, $C$ is set to 1 as follows:
\begin{equation}
\begin{split}
& {\rm Focal}(\widehat{M}, M) \\
& = \frac{-1}{UV} \sum_{uv}\begin{cases}
(1-\widehat{M}_{uv})^\alpha \log (\widehat{M}_{uv}) & {\rm if} \ M_{uv} = 1 \\
(1-M_{uv})^\beta \widehat{M}_{uv}^\alpha \log (1-\widehat{M}_{uv}) & {\rm otherwise}
\end{cases}
\end{split}
\end{equation}
where $U,V$ are the horizon heatmap's size and $u, v$ are the heatmap coordinates. We set focal loss parameters $\alpha = 2, \beta = 4$.

At the post-processing stage, we extract points with the max activation value in its column and then fit a horizon line with the Least Squares Method. 
Such detected 2D horizon line by neural network can be represented as 
\begin{equation}
{\rm HL}^{2d}_{nn}: \quad v = k_h^{nn} u + b_h^{nn} \label{nn PCS horizon line}
\end{equation}
where $u$ and $v$ are pixel coordinates in the image, and $k_h^{nn}$ and $b_h^{nn}$ are slope and intercept of the horizon line detected by neural network, respectively.

\subsubsection{Image Vertical Edge Detection}
In real scenes, the horizon line may be occluded by building or other obstacles, which makes it very hard for neural networks to detect it. Though the vertical edges could assist the horizon line detection and ground plane estimation greatly, there are no annotations of vertical edges in M3OD datasets including KITTI and nuScenes. Considering that the vertical edges are low-level features, we adopt an unsupervised vertical edge slope mining algorithm based on traditional digital image processing technique to implement the vertical line detection.

Our vertical edge slope mining algorithm can be summarized in Algorithm \ref{algor:VerticalEdge}. The input is the original image and the output is the slope of the largest cluster $k_v$ or None.
Firstly, we conduct a GaussianBlur algorithm to the input image, which can exclude tiny textures while not affecting the large color patches.
Secondly, a Canny algorithm is used to extract edge points from the blurred images. Next, a probabilistic Hough transform algorithm is utilized to detect lines.
The GaussianBlur, Canny and HoughLinesP functions are implemented with OpenCV \cite{bradski2000opencv,naveenkumar2015opencv}. Afterwards, we filter the vertical edges with the inclination angle between $80^{\circ}$ and $100^{\circ}$.
Then we employ the Birch \cite{zhang1996birch} algorithm to angles of all vertical edges and find the largest cluster. In this process, noise vertical edges in the image can be eliminated and the vertical edge slope, denoted as $k_v$, can be derived from the centroid slope of the largest cluster.
If the number of the vertical edges $N_V > 3$ and the standard deviation of the slopes  $S_V<3$ , we can trust the slope result of the vertical edges. Then the horizon line is perpendicular to these vertical lines and its slope:
$
k_h = - \frac{1}{k_v}
$.
If the vertical edges are too few or the standard deviation of their slope is too large, the output of the Algorithm \ref{algor:VerticalEdge} would be nothing. In this case, we give up the vertical edge detection and predict the horizon directly. In summary, the vertical edge enhanced image horizon line ${\rm HL}^{2d}_{v}$ is as follows\footnote{The vertical edges have been filtered and their inclination angles are between $80^{\circ}$ and $100^{\circ}$, so $k_v$ cannot be $0$ and $k_h$ is assigned $0$ if $k_v$ is infinity.}:
\begin{equation}
k_h =\begin{cases}
- \frac{1}{k_v}  & {\rm if} \ N_V > 3 \ {\rm and} \ S_V < 3 \\
k_h^{nn} & {\rm otherwise}\end{cases}, \ b_h = b_h^{nn} 
\end{equation}
\begin{equation}
{\rm HL}^{2d}_{v}: \quad v = k_h u + b_h \label{PCS horizon line}
\end{equation}
where $u$ and $v$ are pixel coordinates in images. $k_h^{nn}$ and $b_h^{nn}$ are the slope and intercept of the horizon line detected by neural network, respectively, which can be derived by Eq. \ref{nn PCS horizon line}.
Note that the slope $k_h$ is derived from the vertical edges in the case $N_V>3$ and $S_V<3$ and the intercept $b_h = b_h^{nn}$ is derived from the horizon line detection network, whether we employ the vertical edge or not.
\begin{algorithm}[t]
\caption{Vertical Edge Slope Mining} 
\label{algor:VerticalEdge}
\hspace*{0.02in} {\bf Input:} 
an RGB image, denoted as $img$\\
\hspace*{0.02in} {\bf Output:}
the slope of the largest cluster, denoted as $k_v$, or None
\begin{algorithmic}[1]
\State BlurImg = GaussianBlur($img$, ksize = (13, 13), $\sigma_X$ = 4, 
        $\sigma_Y$ = 4)
\State CannyEdges = Canny(BlurImg, threshold1 = 50, threshold2 = 100, apertureSize=3)
\State HoughEdges = HoughLinesP(CannyEdges, $\rho$ = 1, $\theta$ = $\pi$ / 180, threshold = 5,                     minLineLength=40, maxLineGap=10)
\State VerticalEdgesSlopes = FilterVertical(HoughEdges)
\State $N_V$ = VerticalEdgesSlopes.number
\State $S_V$ = VerticalEdgesSlopes.StandardDeviation
\State LargestCluster = Birch(VerticalEdgesSlopes)
\State $k_v$ = LargestCluster.slope
\If {$N_V>3$ \textbf{and} $S_V<3$}
    \State {\Return $k_v$}
\Else
    \State \Return \textbf{None}
\EndIf
\end{algorithmic}
\end{algorithm}

\subsubsection{Ground Plane Estimation}
\label{section: Ground Plane Estimation}
We aim to acquire the ground plane equation in the camera coordinate from the horizon line in the image. Given the camera intrinsic matrix $\mathbf{K}$, we get the relationship between pixel coordinates $(u,v)^\top$ and the camera coordinates $(x^{c}, y^{c}, z^{c})^\top$ as follows:
\begin{equation}
\begin{aligned}
z_c [u, v, 1]^\top = \mathbf{K}[x^{c}, y^{c}, z^{c}]^\top \  \\
\mathbf{K} = \begin{bmatrix} f_{x} & 0 & c_{u} \\ 0 & f_{y} & c_{v} \\ 0 & 0 & 1 \end{bmatrix} \quad \quad \ \label{K1}
\end{aligned}
\end{equation}
From Eq. \ref{K1}, we can obtain:
\begin{equation}
u = \frac{f_x}{z^{c}} x^{c} + c_u , \quad  v = \frac{f_y}{z^{c}} y^{c} + c_v \label{uv}    
\end{equation}

Considering a pixel $(u,v)^\top$ in the horizon line ${\rm HL}^{2d}_{v}$, we insert Eq. \ref{uv} into Eq. \ref{PCS horizon line}:
\begin{equation}
\frac{f_y}{z^{c}} y^{c} + c_v = k_h (\frac{f_x}{z^{c}} x^{c} + c_u) + b_h \label{xy1}
\end{equation}
In the image, the horizon line's depth value $z^c$ in the CCS is equal to the focal length $f$ of the camera. Therefore, with Eq. \ref{xy1} and $z^c=f$, the horizon line equation in the CCS HL$^{3d}$ can be derived as:
\begin{equation}
{\rm HL}^{3d}: \quad y^c = \frac{k_h f_x}{f_y} x^c + \frac{k_h c_u+ b_h - c_v}{f_y} z^c, \quad z^c = f
\label {HLCCS}
\end{equation}
which represents the image projection of the ground plane when the depth approaches infinity. Based on HL$^{3d}$, we discover an ingenious corollary: the plane (denoted as ${\rm PL}^{oh}$) passing through the horizon line HL$^{3d}$ and the camera’s optical center is parallel to the ground plane (denoted as ${\rm PL}^{gp}$). In the CCS, the intercept between two planes along the Y-axis is the camera's height to the ground, denoted as $H$, and $H$ is nearly invariant since the camera is firmly fixed on the vehicle ($H=1.65m$ in KITTI \cite{KITTI}).
Considering that a plane can be uniquely determined by a line and a point in 3D space, we can derive the equation of plane ${\rm PL}^{oh}$ from the horizon line HL$^{3d}$ (Eq. \ref{HLCCS}) and the camera’s optical center (the origin point of the CCS) as followings:
\begin{align}
{\rm PL}^{oh}: \quad y = \frac{k_h f_x}{f_y} x + \frac{k_h c_u+ b_h - c_v}{f_y} z
\label {P1}
\end{align}
By translating the plane ${\rm PL}^{oh}$ along the Y-axis for $H$, the real-world ground plane ${\rm PL}^{gp}$ can be derived as:
\begin{align}
{\rm PL}^{gp}: \quad y = \frac{k_h f_x}{f_y} x + \frac{k_h c_u+ b_h - c_v}{f_y} z + H
\label {P2}
\end{align}

Additionally, the roll angle \(\theta_{roll}\) and pitch angle \(\theta_{pitch}\) of the ground plane in the CCS can be derived as:
\begin{equation}
\begin{aligned}
\ \ \theta_{roll} = \arctan({\frac{k_h f_x}{f_y}}) \quad \quad \ \\
\theta_{pitch} =\arctan({\frac{k_h c_u +b_h - c_v }{f_y}})
\end{aligned}
\label{egopose}
\end{equation}
\subsubsection{Horizon Line Pseudo Label Generation} 
The horizon line ground truth is indispensable for training the horizon line detection network. However, the wide-used dataset for M3OD, e.g., KITTI \cite{KITTI} and nuSences \cite{caesar2020nuscenes}, does not have the horizon line label annotations. 
Here, we simply use M3OD dataset annotations to generate pseudo labels for horizon line detection, which is cost-efficient and easy-to-operate.

We derive the ground plane pseudo label from the object's 3D bounding box ground truth. Specifically, we use bottom centers of the objects' 3D bounding boxes to fit a plane, which can be regarded as the ground plane in the scene. 
The horizon line pseudo label in the image can be obtained by projecting the infinite position of the ground plane onto the image, which is the inverse process of \nameref{section: Ground Plane Estimation}, seeing Section \ref{section: Ground Plane Estimation}.
\subsection{3D Bounding Box Deduction Based on Dynamic Back Projection}
\label{sec:3D Bounding Box Deduction Based on Dynamic Back Projection}
\subsubsection{Analysis}
Each pixel in the image can be mapped to a ray in space. If we know a point along the ray is on the specific ground, we can determine its spatial position. Note that although the ground plane is not fixed in the CCS, all existing works \cite{groundaware} use a preset ground plane, which lead to a large deviation. Therefore, we design the following dynamic back projection algorithm with our estimated precise contact points in Section \ref{sec:2D Object Localization Guided by Contact Points} and dynamic ground plane equation in Section \ref{sec:Ground Plane Estimation Based on Horizon Line}, then we can derive the accurate 3D BBox.
\subsubsection{Dynamic Back Projection of Contact Points With Ground Plane Equation}
\label{sec:Dynamic Back Projection of Contact Points With Ground Plane Equation}
Given the detected ground contact point $(u_{CP},v_{CP})^\top$ in the image (from Section \ref{sec:Contact Points Detection}) and the camera intrinsic parameter matrix, we can back project the contact point to the spatial position in the CCS:
\begin{equation}
z^c_{CP} [u_{CP}, v_{CP}, 1]^\top = \begin{bmatrix} f_{x} & 0 & c_{u} \\ 0 & f_{y} & c_{v} \\ 0 & 0 & 1 \end{bmatrix} [x_{CP}^{c}, y_{CP}^{c}, z_{CP}^{c}]^\top \label{GCPP} 
\end{equation}
Since the contact point $(x_{CP}^c,y_{CP}^c, z_{CP}^c)^\top$ in the CCS is located on the ground plane, we can conveniently re-formulate Eq. \ref{P2} as follows:
\begin{equation}
y_{CP}^{c} = \frac{k_h f_x}{f_y} x_{CP}^{c} + \frac{k_h c_u+ b_h - c_v}{f_y} z_{CP}^{c} + H \label{GCPGP}
\end{equation}

By solving Eq. \ref{GCPP} and Eq. \ref{GCPGP}, the contact point $(x_{CP}^c,y_{CP}^c, z_{CP}^c)^\top$ in the CCS can be derived using the image coordinate $(u_{CP},v_{CP})^\top$, the horizon line parameters $k_h, b_h$, the camera's height $H$ and the camera's intrinsic parameters as follows:
\begin{equation}
\begin{aligned}
x_{CP}^c = \frac{u_{CP}-c_u}{\lambda} \cdot \frac{f_y}{f_x}\\ \ y_{CP}^c = \frac{v_{CP}-c_v}{\lambda}\ \ \  \\ \  z_{CP}^c = \frac{f_y}{\lambda} \quad \quad \ \ 
\label{GCPxyz0}
\end{aligned}
\end{equation}
where $\lambda = (v_{CP}-k_h u_{CP} -b_h)/ H$.

\subsubsection{3D Attribute Derivation} 
\label{sec:3D Attribute Derivation}
In this section, we will introduce the process of deriving all 3D attributes of an object, i.e., depth, dimension and rotation. Taking Car category as an example, a car has four ground contact points with spatial coordinates $P_{LF}, P_{RF}, P_{RR},$ and $P_{LR}$, derived by Eq. \ref{GCPxyz0} ($LF$, $RF$, $RR$ and $LR$ represent the left-front, right-front, right-rear and left-rear point, respectively.).
Then the estimated bottom center coordinate can be calculated as: 
\begin{equation}
    P_{BC} = \frac{1}{4} (P_{LF}+ P_{RF}+ P_{LR}+ P_{RR}) \label{PBC}
\end{equation}
The depth of the car $d_g$ from ground plane prior can be derived as:
\begin{equation}
    d_g = z_{BC} = \frac{1}{4} (z_{LF}+ z_{RF}+ z_{LR}+ z_{RR}) \label{dgzbc}
\end{equation}
where $ z_{BC}, z_{LF}, z_{RF}, z_{LR}, z_{RR}$ are the Z-axis coordinates of $P_{BC}, P_{LF}, P_{RF}, P_{LR},$ $P_{RR}$ in the CCS, respectively.

In the process of contact point pseudo label generation (Section \ref{sec:Contact Point Pseudo Label Generation}), we have defined $k_l$ as the ratio of the front and rear wheels' distance to the 3D BBox's length and $k_w$ as the ratio of the left and right wheels' distance to the 3D BBox's width. Then the length $l_{3D}$ and the width $w_{3D}$ of the 3D BBox can be derived from the wheels' distance:
\begin{equation}
\begin{aligned}
& l_{3D} = \frac{\Vert(P_{LF}+ P_{RF})- ( P_{LR}+ P_{RR})\Vert_2}{2 k_l}   \\
& w_{3D} = \frac{\Vert(P_{RF}+ P_{RR})- ( P_{LF}+ P_{LR})\Vert_2 }{2 k_w} \label{carlw}
\end{aligned}
\end{equation}
The height of the 3D BBox
is $h_{3D} = d_g \cdot h_{2D}/ f_y$, where $f_y$ is one of camera intrinsic parameters, $d_g$ is the depth of the object from Eq. \ref{dgzbc} and $h_{2D}$ is the predicted object's 2D BBox height by neural network.
\footnote{For other categories which have less than four contact points, we use the average width and length to estimate the geometry based $w_{3D}$ and $l_{3D}$. Apart from that, the process of acquiring 3D attributes follows a similar manner to that of Car category.}

The rotation angle of the object in the BEV (X-Z plane) from the X-axis is as following:
\begin{equation}
r_o = \arctan({\frac{z_{LF}+z_{RF}-2z_{BC}}{x_{LF}+x_{RF}-2x_{BC}}}), \ \ r_o \in [-\frac{\pi}{2}, \frac{\pi}{2}]
\label{r0eqn}
\end{equation}
The final rotation result $r_g$ is $r_0$ after simple adjustments according to the car's orientation, i.e., $r_g=r_0$, $r_g=r_0+\pi$ or $r_g=r_0-\pi$, where $r_0 \in [-\frac{\pi}{2}, \frac{\pi}{2}]$ and $r_g \in [-\pi,\pi]$.

From the above procedure, we can derive all required 3D attributes for M3OD, i.e., depth, dimension ($l_{3D}$, $w_{3D}$ and $h_{3D}$) and rotation. Note that in most M3OD works, the dimension and rotation of the object are usually predicted by neural networks directly and could not be derived by geometry deduction. By contrast, the 3D BBox could be determined more accurately in our approach through the geometry derivation of the dimension and rotation.

\subsubsection{3D Bounding Box Refinement}
In order to enhance the prediction of 3D BBox, we employ a refining network, dubbed RefineNet, to accomplish 3D BBox refinement. Following previous works~\cite{gupnet,monogrnet}, we use RoIAlign \cite{he2017mask} to extract the single object's RoI feature. 
The RefineNet exploits the RoI feature as input and has independent detection heads for the 3D center offset and the 3D attribute bias. The 3D center offset is the offset of the 3D center's projection in the image and the 3D attribute bias is comprised of rotation bias, 3D size bias and depth bias.
Note that we have derived the geometry estimated depth $d_g$,  rotation $r_g$ and 3D size $s_g = (w_{3D}, h_{3D}, l_{3D})^\top$ in Section \ref{sec:3D Attribute Derivation}. Then the refined depth $\widehat{d}$, rotation $\widehat{r}$ and 3D size $\widehat{s}$ can be obtained by adding the geometry estimation $d_g, r_g, s_g$ to the network output bias $d_b, r_b, s_b$, respectively.
\begin{equation}
\begin{split}
 \widehat{d} = d_g + d_b, \ \widehat{s} = s_g + s_b, \ \widehat{r} = r_g + r_b 
 \end{split} \label{b+p1}
\end{equation}

Following GUPNet~\cite{gupnet}, we employ the uncertainty mechanism and model the depth $\widehat{d}$ as a Laplace distribution, and its standard deviation is $\sigma$.
Finally, the depth, size and rotation losses are as follows:
\begin{equation}
\begin{aligned}
L_{depth} = 
    \frac{\sqrt{2}}{\sigma}\left|\widehat{d}-d\right|+\log(\sigma) \ \ \ \\
L_{dim} =  {\rm L_1} (\widehat{s}, s), \ 
L_{rot} = {\rm L_1} (\widehat{r}, r) \ 
\end{aligned}
\end{equation}
where $d$, $s$ and $r$ are the corresponding ground truths. As for the 3D center offset, we define $\widehat{\delta}_{3d}$ and $\delta_{3d}$ as the predicted value and its corresponding ground truth, then the 3D center offset loss $L_{off3d}$ is: 
\begin{equation}
    L_{off3d} = {\rm L_1} (\widehat{\delta}_{3d}, \delta_{3d})
\end{equation}
The total loss $L_{total}$ is a weighted sum of all tasks according to Eq. \ref{eq:loss}. In our experiment, the 2D weight $w_{2d}$ is set to 0.1 and the other weights are set to 1.0.
\begin{equation}
\begin{split}
L_{total}  = & w_{kps} L_{kps} + w_{2d} L_{2d} + w_{hor} L_{hor} +w_{dim} L_{dim} \\
& +w_{rot} L_{rot}+w_{depth} L_{depth} +w_{off3d} L_{off3d}
\end{split}
\label{eq:loss}
\end{equation}

\begin{table*}[htbp]
\begin{center}
    \caption{{\bf 3D and BEV AP$_{40}$ scores on the KITTI 3D object detection \emph{test} set for Car category at IoU 0.7.} We highlight the best results in bold and underline the second best results.} 
    \label{table:kitti_test_3DBEV}
    \centering
    
 \begin{tabular}{c||c|c|c c c|c c c}
	\toprule[1pt]
    \multirow{2}{*}{Method} & \multirow{2}{*}{Year}    & \multirow{2}{*}{Extra data}   & \multicolumn{3}{c|}{AP$_{{\rm 3D}}$}                     & \multicolumn{3}{c}{AP$_{{\rm BEV}}$} \\ \cline{4-9} 
    &&& Easy & Moderate & Hard & Easy & Moderate & Hard \\ \hline
    MonoPSR~\cite{ku2019monocular}                      & 2019 & LiDAR                & 10.76  & 7.25  & 5.85  & 18.33  & 12.58  & 9.91 \\
    AM3D~\cite{ma2019accurate}                       & 2019  & Depth       & 16.50  & 10.74 & 9.52  & 25.03  & 17.32  & 14.91 \\
    Decoupled-3D~\cite{cai2020monocular}               & 2020  & Depth               & 11.08  & 7.02  & 5.63  & 23.16  & 14.82  & 11.25 \\
    PatchNet~\cite{ma2020rethinking}                   & 2020  & Depth         & 15.68  & 11.12 & 10.17 & 22.97  & 16.86  & 14.97 \\
    DA-3Ddet~\cite{ye2020monocular}                   & 2020  & Depth        & 16.80  & 11.50 & 8.90 & -  & -  & - \\
    D4LCN~\cite{ding2020learning}                      & 2020  & Depth               & 16.65  & 11.72 & 9.51  & 22.51  & 16.02  & 12.55 \\
    Kinem3D~\cite{brazil2020kinematic}                    & 2020  & Multi-frames      & 19.07  & 12.72 & 9.17  & 26.69  & 17.52  & 13.10 \\
    CaDDN~\cite{reading2021categorical}                      & 2021  & LiDAR              & 19.17  & 13.41 & 11.46 & 27.94  & 18.91  & 17.19 \\
    DFR-Net~\cite{zou2021devil}                       & 2021  & Depth                & 19.40  & 13.63 & 10.35  & 28.17  & 19.17  & 14.84 \\
    MonoEF~\cite{zhou2021monocular}                     & 2021  & External                 & 21.29  & 13.87 & 11.71 & 29.03  & 19.70  & 17.26 \\
     \hline
    MonoDIS \cite{simonelli2019disentangling}         & 2019 & None                 & 10.37  & 7.94  & 6.40  & 17.23  & 13.19  & 11.12 \\
    M3D-RPN~\cite{brazil2019m3d}                      & 2019 & None                 & 14.76  & 9.71  & 7.42  & 21.02  & 13.67  & 10.23 \\
    SMOKE~\cite{smoke}                      & 2020 & None           &  14.03  & 9.76  & 7.84  & 20.83  & 14.49  & 12.75 \\
    MonoPair~\cite{chen2020monocular}                  & 2020  & None                & 13.04  & 9.99  & 8.65  & 19.28  & 14.83  & 12.89 \\
    MonoRCNN~\cite{shi2021geometry}                 & 2021  & None           &  18.36  & 12.65 & 10.03  & 25.48  & 18.11  & 14.10 \\
    DDMP-3D~\cite{wang2021depth}                    & 2021  & None           &  19.71  & 12.78 & 9.80  & 28.08  & 17.89  & 13.44 \\
    MonoFlex~\cite{zhang2021objects}                   & 2021  & None                & 19.94  & 13.89 & 12.07 & 28.23  & 19.75  & 16.89 \\ 
    Fine-Grained~\cite{liu2022fine}                   & 2022  & None                &  20.28  & 13.12 & 9.56 & - & -  & - \\
    MonoGround~\cite{qin2022monoground}                   & 2022  & None                &  21.37  & 14.36 & 12.62 & \underline{30.07} & \underline{20.47}  & \underline{17.74}\\
    MonoDTR~\cite{huang2022monodtr}                   & 2022  & None                &  \underline{21.99}  & \underline{15.39} & \underline{12.73} & 28.59 & 20.38  & 17.14\\
    \hline
    GPENet (Ours)                 & 2022   & None                 & \textbf{22.41}  & \textbf{15.44} & \textbf{12.84} & \textbf{30.31}  & \textbf{20.79}  & \textbf{18.21} \\ \bottomrule
    \end{tabular}
    \end{center}
\end{table*}

{
\begin{table*}[htbp]
\begin{center}
\caption{{\bf AP$_{40}$ scores on the KITTI 3D object detection \emph{test} set for Car, Pedestrian and Cyclist category.} We highlight the best results in bold and underline the second best results.}
\label{table:test}
\begin{tabular}{c||c|c c c|c c c|c c c}
	\toprule[1pt]
			\multirow{2}{*}{Method} & \multirow{2}{*}{Extra data} & 
			\multicolumn{3}{c|}{Car@IoU=0.7} & \multicolumn{3}{c|}{Pedestrian@IoU=0.5} & \multicolumn{3}{c}{Cyclist@IoU=0.5} \\ \cline{3-11} 
		& & Easy & Mod.& Hard & Easy & Mod. & Hard & Easy & Mod. & Hard\\ \hline
		D4LCN \cite{ding2020learning} & Depth & 16.65 & 11.72 &9.51 &4.55 &3.42& 2.83 &2.45 &1.67 &1.36 \\
		Kinematic \cite{brazil2020kinematic} & Multi-frames &19.07&12.72&9.17 &-&-& -&-&-&- \\
		MonoEF \cite{zhou2021monocular} & External &  21.29 &	13.87 &	11.71
        &4.27 &	2.79 &	2.21 &1.80 &	0.92 &	0.71\\
		\hline
		MonoPair \cite{chen2020monocular} & None &13.04&9.99&8.65&10.02 & 6.68 & 5.53 & 3.79 & 2.12 & 1.83 \\ 
		M3DSSD \cite{Luo_2021_CVPR} & None & 17.51&11.46&8.98&5.16& 3.87&3.08&2.10& 1.51&1.58 \\
		MonoDLE \cite{ma2021delving} & None  &17.23&12.26&10.29&5.34&3.28&2.83&4.59&2.66&2.45 \\
		GrooMeD-NMS \cite{kumar2021groomed} & None & 18.10&12.32&9.65&-&-&-&-&-&-\\
		MonoFlex \cite{zhang2021objects} & None & 19.94 & 13.89 & 12.07 & 9.43 & 6.31 & 5.26 & 4.17 & 2.35 & 2.04\\
		GUPNet \cite{gupnet} & None &20.11&14.20&11.77&{\bf 14.72}&{\bf 9.53}& \underline{7.87}&4.18&2.65&2.09\\
		MonoGround \cite{qin2022monoground} & None &\underline{21.37}&\underline{14.36}&\underline{12.62}&12.37&7.89& 7.13&\underline{4.62}&\underline{2.68}&\underline{2.53}\\ \hline
		GPENet (Ours) & None &{\bf22.41}&{\bf15.44}&{\bf12.84}&\underline{14.61}&\underline{9.36}&{\bf7.91}&{\bf 5.45}&{\bf 3.05}&{\bf 2.56} \\
		\bottomrule[1pt]
	\end{tabular}
\end{center}
\end{table*}
\setlength{\tabcolsep}{1.4pt}}

\section{Experiments}
\subsection{Setup and Implementation Details}
We conduct experiments on the KITTI \cite{KITTI} and nuScenes \cite{caesar2020nuscenes} dataset to verify the effectiveness and generalization of the proposed method. The KITTI 3D dataset is the most popular benchmark in M3OD. Following~\cite{chen2016monocular,chenmulti}, we split the training images into training and validation set and use the standard AP$_{40}$ and AP$_{11}$ \cite{KITTI, simonelli2019disentangling} to evaluate our method. Both the M3OD and bird's-eye view detection tasks based on the standard evaluation protocol are taken into consideration. Besides, we also conduct experiments on the nuScenes \cite{caesar2020nuscenes} to demonstrate the generalization capacity of our approach.

Our code is implemented with PyTorch 1.7.1 \cite{paszke2019pytorch} and CUDA 11.0. The input images are resized to $1280 \times 384$. We utilize the popular DLA34\cite{dla} as the backbone with a downsample ratio of $4$.  For all detection heads, we use two $3 \times 3$ convolution layers with batch normalization and ReLU between them. We train our network with the batch size 16 and Adam~\cite{kingma2014adam} optimizer for 200 epochs. The initial learning rate is $1.25 \times 10^{-3}$, and we use the cosine learning rate warmup schedule of 5 epochs and learning rate decay of 0.1 in the training process.

\subsection{Main Results}
\subsubsection{Results on the KITTI Test Set}
Table \ref{table:kitti_test_3DBEV} is 3D and BEV results for Car category at IoU 0.7 on the KITTI test set. The results are evaluated on KITTI's official test server. For the most important Car category, our method achieves the best performance in all six metrics of the 3D and BEV detection without any extra data. In AP$_{3D}$ for car, our GPENet has achieved 22.41\%, 15.44\% and 12.84\% in the easy, moderate and hard levels, respectively. In AP${_{{\rm BEV}}}$, we have achieved 30.31\%, 20.79\% and 18.21\% in three levels. As we can see, our approach outperforms previous state-of-the-art methods in both 3D and BEV metrics. Moreover, Table \ref{table:test} shows our results of all three categories on the KITTI test set, i.e., Car, Pedestrian and Cyclist. Our method has achieved the state-of-the-art performance for Car and Cyclist category and performs excellently for Pedestrian category.

Note that some other works \cite{ku2019monocular, ma2019accurate, cai2020monocular, brazil2020kinematic} use extra data during the training stage or the inference stage, such as LiDAR points, depth map and external data, which requires much higher annotation costs and makes neural networks even more complex.

\subsubsection{AP$_{40}$ and AP$_{11}$ Results of Car Category on the KITTI Validation Set}
As indicated in Table \ref{table:validation40} and Table \ref{table:validation11}, we use the KITTI official metrics AP$_{40}$ and AP$_{11}$ \cite{KITTI} to evaluate GPENet's performance on the validation set \cite{chen2016monocular}.

Our method ranks top in various metrics, demonstrating that our approach does work and yields better results. 
Especially, our approach is far superior to the competitors when the IoU is 0.5.

In order to fully exploit the great power of ground plane prior, we use a manually labeled contact point training set to obtain better contact point detection results and achieve more accurate 2D-3D reasoning. We use GPENet* to represent this setup. As shown in the last line of Table \ref{table:validation40} and Table \ref{table:validation11}, our GPENet* further improves the results of GPENet in most metrics and both GPENet and GPENet* achieve state-of-the-art performance.
This result further demonstrates the effectiveness of our proposed method and we hope our work can provide a new perspective to pick up the ground plane prior for M3OD community.

\setlength{\tabcolsep}{3pt}
\begin{table*}[htbp]
\begin{center}
\caption{{\bf AP$_{40}$ scores on the KITTI 3D object detection \emph{validation} set for Car category.} We highlight the best results in bold.}
\label{table:validation40}

\begin{tabular}{c||c c c|c c c|c c c|c c c}
	\toprule[1pt]
			\multirow{2}{*}{Method} & 
			\multicolumn{3}{c|}{3D@IoU=0.7} &
			\multicolumn{3}{c|}{BEV@IoU=0.7} &
			\multicolumn{3}{c|}{3D@IoU=0.5} & \multicolumn{3}{c}{BEV@IoU=0.5} \\ \cline{2-13} 
		& Easy & Mod.& Hard & Easy & Mod. & Hard & Easy & Mod. & Hard & Easy & Mod. & Hard\\ \hline
		MonoGRNet   \cite{monogrnet}				  		&11.90&7.56&5.76&19.72&12.81&10.15&47.59&32.28&25.50&48.53&35.94&28.59\\ 
		MonoDIS     \cite{simonelli2019disentangling}	&11.06&7.60&6.37&18.45&12.58&10.66&-&-&-&-&-&-\\ 
		M3D-RPN     \cite{brazil2019m3d}					&14.53&11.07&8.65&20.85&15.62&11.88&48.53&35.94&28.59&53.35&39.60&31.76\\ 
		MoVi-3D     \cite{simonelli2020towards}			&14.28&11.13&9.68&22.36&17.87&15.73&-&-&-&-&-&-\\ 
		MonoPair    \cite{chen2020monocular}				&16.28&12.30&10.42&24.12&18.17&15.76&55.38&42.39&37.99&61.06&47.63&41.92\\
		MonoDLE     \cite{ma2021delving} 				&17.45&13.66&11.68&24.97&19.33&17.01&55.41&43.42&37.81&60.73&46.87&41.89 \\
		MonoFENet     \cite{ma2021delving} 				&17.54&11.16&9.74&30.21&20.47&17.58&59.93&42.67&37.50&66.43&47.96&43.73 \\
		GrooMeD-NMS \cite{kumar2021groomed} 				&19.67&14.32&11.27&27.38&19.75&15.92&55.62&41.07&32.89&61.83&44.98&36.29 \\
		MonoFlex    \cite{zhang2021objects}				&23.64&17.51&14.83&-&-&-&-&-&-&-&-&-\\
		GUPNet      \cite{gupnet}  						&22.76&16.46&13.72&31.07&22.94&19.75&57.62&42.33&37.59&61.78&47.06&40.88\\ 
		\hline
		GPENet (Ours)  &23.96&17.35&14.40&32.31&23.68& 20.32&62.89&{\bf 47.56}&{\bf 42.48}&68.02&{\bf 52.30}&{\bf 47.24}\\
		GPENet* (Ours) & {\bf 25.47}  & {\bf 17.80} & {\bf 14.84} &{\bf 33.50}&{\bf 24.04}&{\bf 20.68}&{\bf 63.85}&47.45&41.82& {\bf 68.90}& 51.35& 46.13\\
		\bottomrule[1pt]
	\end{tabular}
	\vspace{-0.2cm}
\end{center}
\end{table*}
\setlength{\tabcolsep}{1.4pt}

\subsubsection{The Depth and Dimensions $L_1$ Errors on the KITTI Validation Set}
We evaluate the $\rm L_1$ errors of the most important 3D attributes in M3OD, including object depth, 3D height, 3D length and 3D width on the KITTI validation set, as shown in Table \ref{table:depth and dimensions}. Improvement over the previous state-of-the-art method \cite{gupnet} demonstrates the outstanding performance of our approach.

\setlength{\tabcolsep}{4pt}
\begin{table*}[htbp]
\begin{center}
\caption{{\bf AP$_{11}$ scores on the KITTI 3D object detection \emph{validation} set for Car category.} We highlight the best results in bold.}
\label{table:validation11}

\begin{tabular}{c||c c c|c c c|c c c|c c c}
	\toprule[1pt]
			\multirow{2}{*}{Method} & 
			\multicolumn{3}{c|}{3D@IoU=0.7} &
			\multicolumn{3}{c|}{BEV@IoU=0.7} &
			\multicolumn{3}{c|}{3D@IoU=0.5} & \multicolumn{3}{c}{BEV@IoU=0.5} \\ \cline{2-13} 
		& Easy & Mod.& Hard & Easy & Mod. & Hard & Easy & Mod. & Hard & Easy & Mod. & Hard\\ \hline
	    Mono3D \cite{chen2016monocular}  
	    &2.53 &2.31 &2.31 &5.22 &5.19 &4.13 &-&-&-&-&-&-\\
	    Deep3DBox  \cite{mousavian20173d}
	    &5.85 &4.19 &3.84 &9.99 &7.71 &5.30 &27.04 &20.55 &15.88 &30.02 &23.77 &18.83\\
	    Mono3D++ \cite{he2019mono3d}
	    &10.60 &7.90& 5.70 &16.70 &11.50 &10.10 &42.00 &29.80 &24.20 &46.70 &34.30 &28.10 \\
	    M3D-RPN \cite{brazil2019m3d}
	    &20.27 &17.06 &15.21 &25.94 &21.18 &21.18 &48.96 &39.57 &33.01 &53.35 &39.60 &31.76 \\
	    RTM3D \cite{li2020rtm3d}
	    &20.77 &20.77 &16.63 &25.56 &22.12 &20.91 &54.36 &41.90 &35.84 &57.47 &44.16 &42.31 \\
	    RARNet \cite{liu2020reinforced}+M3D-RPN\cite{brazil2019m3d}
	    &23.12 &19.82 &16.19 &29.16 &22.14 &18.78 &51.20 &44.12 &32.12 &57.12 &44.41 &37.12 \\
		GUPNet      \cite{gupnet}  						
		&25.76 & 20.48 & 17.24 & 34.00 & 24.81 & 22.96 & 59.36 & 45.03 & 38.14 & 62.58 & 46.82 & 44.81\\ 
		\hline
		GPENet (Ours)  &27.72 & 22.79 & 19.12 & 35.34 & {\bf 27.09} & {\bf 25.05 } & 63.52 & {\bf 47.82 } & {\bf 44.87 } & 66.50 & {\bf 54.17} & {\bf 47.74 }\\
		GPENet*(Ours) & {\bf 28.85}  & {\bf 22.90} & {\bf 19.41} &{\bf 36.30} & 26.90 & 24.98 & {\bf 63.71 } & 47.54 & 44.77 & {\bf 66.85} & 54.05 & 47.03\\
		\bottomrule[1pt]
	\end{tabular}
	\vspace{-0.3cm}
\end{center}
\end{table*}
\setlength{\tabcolsep}{1.4pt}

{\setlength{\tabcolsep}{4pt}
\begin{table}
\renewcommand\arraystretch{1.2}
\begin{center}
\caption{The depth and dimensions $\rm L_1$ errors for KITTI validation set (meters). The lower is the better.}
\label{table:depth and dimensions}
\begin{tabular}{c|c c c c }
	\toprule[1pt]
			Method & object depth & 3D height & 3D length & 3D width \\ \hline
		GUPNet \cite{gupnet} & 1.362 & 0.084 & 0.324 & 0.087 \\
		GPENet (Ours) & 0.991 &0.073 &0.288 &0.075\\
		\bottomrule[1pt]
	\end{tabular}
\vspace{-0.2cm}
\end{center}
\end{table}

\setlength{\tabcolsep}{1.4pt}}

{\setlength{\tabcolsep}{2pt}
\begin{table}
\renewcommand\arraystretch{1.2}
\begin{center}
\caption{{\bf Influence of ground plane estimation on M3OD AP$_{40}$ results for Car category.} Exp (a) is the baseline in which the object's 3D attributes are predicted directly. 
BP here represents the back projection with the ground plane prior.
Horizon here represents the ground plane estimation by direct prediction of the horizon through the network. VerE means the vertical line detection.}
\label{table:ablationGroundPlane}
\begin{tabular}{c|c c c |c c c|c c c}
	\toprule[1pt]
			\multirow{2}{*}{Exp} & \multirow{2}{*}{BP} & \multirow{2}{*}{Horizon} & \multirow{2}{*}{VerE} & \multicolumn{3}{c|}{3D@IoU=0.7} & \multicolumn{3}{c}{BEV@IoU=0.7}  \\ \cline{5-10} 
		& & & & Easy & Mod.& Hard & Easy & Mod. & Hard \\ \hline
		(a)  & - & - & -                           &17.36&12.97&10.74&24.80&18.38&16.33 \\ 
		(b) & \checkmark  & - & -                &17.62&14.04&11.88&24.74&19.12&17.15 \\ 
		(c) & \checkmark & \checkmark & -        &20.33&15.30&13.24&28.52&21.69&18.33 \\ 
		(d) & \checkmark & - & \checkmark 		  &22.71&15.64&13.51&29.52&21.45&18.40 \\ 
		(e) & \checkmark & \checkmark & \checkmark &\textbf{23.96}&\textbf{17.35}&\textbf{14.40}&\textbf{32.31}&\textbf{23.68}&\textbf{20.32} \\ 
		\bottomrule[1pt]
	\end{tabular}
\vspace{-0.5cm}
\end{center}
\end{table}
\setlength{\tabcolsep}{1.4pt}}
\subsection{Ablation Study}
To verify the effectiveness of our approach's architecture, we conduct a detailed ablation study on the KITTI validation set, as shown in Table~\ref{table:ablationGroundPlane},\ref{table:Results of different detection points}. 
\subsubsection{Effectiveness of the Ground Plane Estimation} In Table~\ref{table:ablationGroundPlane}, Exp (a) is the baseline in which the object's 3D attributes are predicted directly. Exp (b) uses the preset fixed ground plane to back project the contact points and deduce the 3D BBox. By comparing the experimental results (a $\rightarrow$ b), we discover that the introduction of the back projection with ground plane prior improves accuracy marginally. This observation proves our motivation that directly using the preset fixed ground plane is likely to encounter the problem of ground plane tilt, thus resulting in non-negligible errors. 
Exp (c) accomplishes ground plane estimation by direct prediction of the horizon line through the network without the vertical edges' assistance.
From Exp (b) to Exp (c), the 3D AP$_{40}$ increases by 1.26\% on the moderate level.
Exp (d) is an interesting setting which uses vertical edge detection in the image to calculate the roll angle of the ground plane and assumes that the pitch angle of the ground plane is zero.
After the vertical edge detection is equipped, the 3D AP$_{40}$ on the moderate setting increases by 1.60\% (b $\rightarrow$ d). Exp (e) is our complete GPENet. It uses vertical-edge-enhanced horizon line detection to estimate the ground plane equation, which makes the 3D AP$_{40}$ increase to 23.96\%, 17.35\%, 14.40\% on the easy, moderate and hard settings, respectively. 
\begin{figure*}[htbp]
	\begin{center}
		\includegraphics[width=492pt]{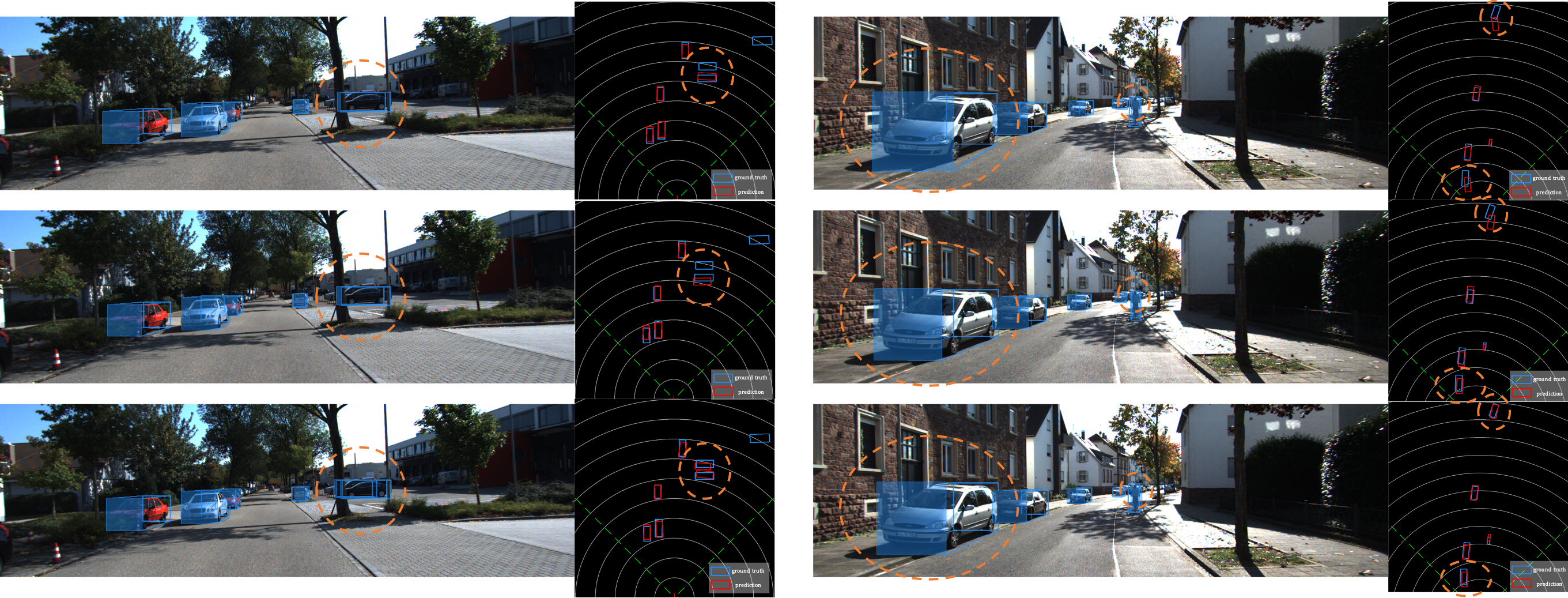}
	\end{center}
	\vspace{-0.2cm}
	\caption{\textbf{Visualization results.} From top to bottom, we show examples generated by using 3D boxes's bottom vertices as projection points, by using a preset fixed ground plane, and by our GEPNet, respectively.}
	\label{fig:visual}
\end{figure*}
{\setlength{\tabcolsep}{4pt}
\begin{table}
\renewcommand\arraystretch{1.2}
\begin{center}
\vspace{-0.3cm}
\caption{{\bf Influence of projection points on M3OD AP$_{40}$ results.} BVB means the bottom vertices of 3D BBox and GCP means the ground contact points.}
\label{table:Results of different detection points}
\begin{tabular}{c|c c c}
	\toprule[1pt]
			\multirow{2}{*}{Keypoint} &  
			\multicolumn{3}{c}{Car@IoU=0.7 \ \ $\rm AP_{3D}$/$\rm AP_{BEV}$} \\ \cline{2-4} 
		& Easy & Mod.& Hard \\ \hline
		BVB & 11.49/16.04 & 8.70/13.27 & 7.72/11.59 \\
		GCP & 23.96/32.31 & 17.35/23.68& 14.40/20.32\\
		\bottomrule[1pt]
	\end{tabular}
	
	\begin{tabular}{c|c c c}
	\end{tabular}
	
	\begin{tabular}{c|c c c}
	\end{tabular}
	
	\begin{tabular}{c|c c c}
	\toprule[1pt]
			\multirow{2}{*}{Keypoint} &  
	 \multicolumn{3}{c}{Cyclist@IoU=0.5 \ \ $\rm AP_{3D}$/$\rm AP_{BEV}$} \\ \cline{2-4} 
		& Easy & Mod.& Hard \\ \hline
		BVB & 7.84/7.07 & 2.44/3.84 & 1.95/3.56\\
		GCP & 7.85/9.09 & 4.48/5.53 & 3.97/4.74 \\
		\bottomrule[1pt]
	\end{tabular}
\end{center}
\vspace{-0.4cm}
\end{table}
\setlength{\tabcolsep}{1.4pt}}
\subsubsection{Influence of the Ground Contact Points} As shown in Table \ref{table:Results of different detection points}, we investigate the improvement of using ground contact points (GCP) instead of the bottom vertices of objects' 3D bounding box (BVB). We use our estimated accurate ground plane equation in dynamic back projection for both GCP and BVB. From the results, the AP$_{40}$ increases dramatically when using GCP as the projection points, especially for Car category (from 8.70\%/13.27\% to 17.35\%/23.68\% on the moderate level).
According to the result, the precise ground contact point detection is critical to leverage the ground plane reference.
{\setlength{\tabcolsep}{4pt}
\begin{table}
\renewcommand\arraystretch{1.2}
\begin{center}
\vspace{-0.3cm}
\caption{Cross-dataset evaluation on KITTI and nuScenes frontal validation dataset. }
\label{table:Cross-dataset evaluation}
\begin{tabular}{l|l|c c c}
	\toprule[1pt]
	        \multirow{2}{*}{Dataset} &  
			\multirow{2}{*}{Method} &  
			\multicolumn{3}{c}{Depth~prediction~mean~error~(meters)~\textdownarrow} \\ \cline{3-5} 
		& & $[0, 20)$\ \ \ \ \ \  & $[20, 40)$ & $[40, +\infty)$ \\ \hline
\multirow{4}{*}{KITTI}& M3D-RPN\cite{brazil2019m3d} & 0.56\ \ \ \ \ \   & 1.33  & 2.73 \\
& MonoRCNN\cite{shi2021geometry}  & \textbf{0.46}\ \ \ \ \ \   & 1.27  & 2.59 \\
& GUPNet \cite{gupnet} & 0.54\ \ \ \ \ \  & 1.21& 2.49 \\
& GPENet (Ours)       & \textbf{0.46}\ \ \ \ \ \  & \textbf{1.12} &\textbf{2.22}  \\ \hline
	\multirow{4}{*}{nuScenes}& M3D-RPN\cite{brazil2019m3d} & 1.04\ \ \ \ \ \   & 3.29  & 10.73 \\
& MonoRCNN\cite{shi2021geometry} & 0.94\ \ \ \ \ \   & 2.84   & 8.65 \\
& GUPNet \cite{gupnet} & 0.83\ \ \  \ \ \ & 2.14 &5.98 \\
& GPENet (Ours)       & \textbf{0.72}\ \ \ \ \ \  & \textbf{2.11} &\textbf{5.79} \\ 
		\bottomrule[1pt]
	\end{tabular}
	\vspace{-0.5cm}
\end{center}
\end{table}
\setlength{\tabcolsep}{1.4pt}}
\subsection{Cross-Dataset Evaluation}
To demonstrate the generalization capacity of our approach, we follow \cite{shi2021geometry} and conduct cross-dataset evaluation. As shown in Table \ref{table:Cross-dataset evaluation}, all models are trained on the KITTI train set, evaluated on the KITTI validation set and the nuScenes frontal validation set. Considering that depth prediction is the most critical item in M3OD, we evaluate the depth mean errors in different depth ranges. The results show that our approach outperforms other works, which confirms the generalization capacity of GPENet.

\begin{figure*}[htbp]
	\begin{center}
		\includegraphics[width=511pt]{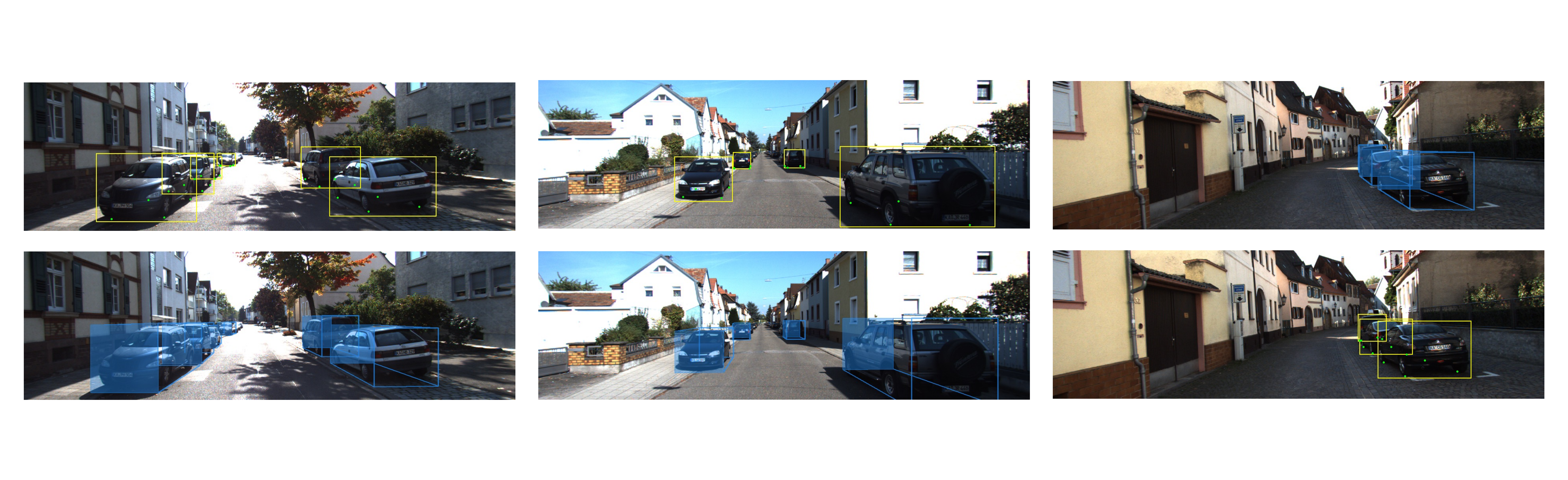}
	\end{center}
		\vspace{-0.2cm}
\caption{\textbf{ Examples of the vertical edge and horizon line detection.} The green lines are detected vertical edges and the red lines represent the detected horizon lines in the image. The proposed vertical-edge-enhanced horizon line detection method can work effectively even if the horizon line is heavily occluded by obstacles.}
\vspace{-0.2cm}
	\label{fig:verhor}
\end{figure*}
\begin{figure*}[htbp]
	\begin{center}
		\includegraphics[width=511pt]{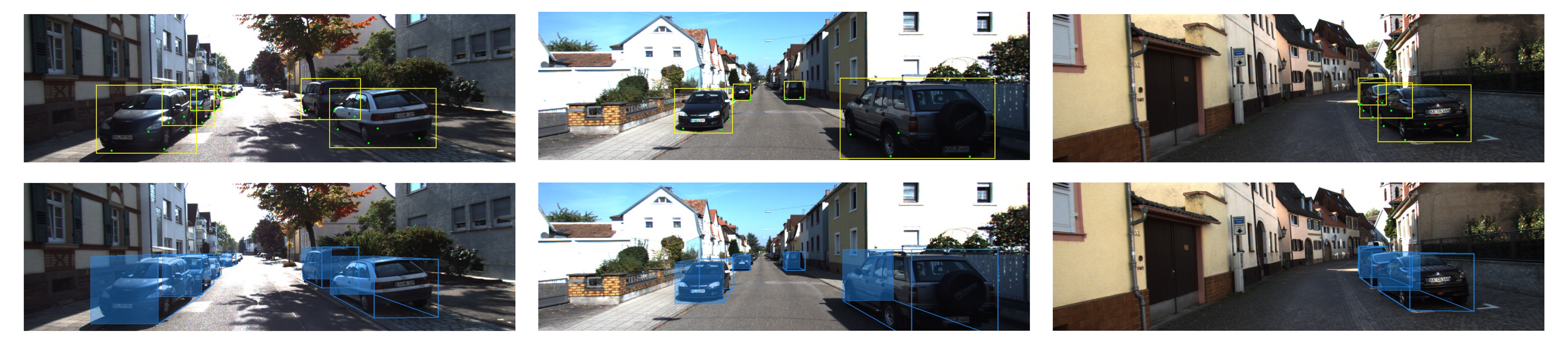}
	\end{center}
	\vspace{-0.25cm}
\caption{\textbf{The 2D BBox, ground contact point and 3D BBox detection results.} They are represented by yellow, green and blue, respectively. Although some contact points are blocked in the image, our model can still work well.}
\vspace{-0.5cm}
	\label{fig:2D_3D}
\end{figure*}

\subsection{Visualization Analysis} We visualize some examples to further demonstrate the effectiveness of the proposed GPENet. As shown in Fig \ref{fig:visual}, when two objects are close to each other, our GPENet can successfully separate them with clear boundary, while both competitors fail. Besides, compared with two competitors, our method tends to predict a more accurate spatial position. For example, in the example on the right, the car circled by the bigger orange ellipse has the best-fitting 3D boxes generated by our GPENet, which reveals a more accurate 3D size estimation. Meanwhile, as shown in bird’s-eye view, the distance between the prediction result and ground truth is apparently smaller, especially for distant objects (circled by the smaller orange ellipse). This observation indicates that the proposed GPENet can produce higher quality 3D bounding boxes. Fig. \ref{fig:verhor} is our vertical edge and horizon line detection result. As we can see, the proposed vertical-edge-enhanced horizon line detection method can work effectively even if the horizon is heavily occluded by obstacles. Fig. \ref{fig:2D_3D} shows the 2D BBox, ground contact points and 3D BBox detection result. Note that our model can still work when some contact points are occluded in the image, which demonstrates the robustness of our approach.

\section{Conclusions}
\vspace{-0.015cm}
In this paper, we identify two key drawbacks that highly hinder the application of the ground plane prior in modern monocular 3D object detection, i.e., the projection point localization issue and the ground plane tilt problem. To tackle both issues, we propose a Ground Plane Enhanced Network, namely GPENet, which aims to pick up the ground plane prior for M3OD community. Specifically, to deal with the projection point localization issue, we adopt the objects’ ground contact points as explicit signals of objects. Such ground contact points can be well projected into the 3D space with the ground plane equation, resulting in a more accurate geometry estimation. Besides, we introduce the horizon line in the image to estimate the ground plane equation, solving the ground plane tilt problem. We also propose a vertical-edge-enhanced method for the horizon line detection when the horizon is highly occluded. A 3D BBox deduction based on the dynamic back projection method is proposed to make use of the contact points and ground plane equation. In our method, pseudo labels for contact points and horizon lines are generated using only M3OD labels. We conduct extensive experiments on the popular KITTI and nuScenes datasets. Experimental results show that the proposed GPENet can significantly outperform baseline methods, obtaining state-of-the-art performance, which demonstrates the effectiveness and superiority of the proposed method.

\bibliographystyle{IEEEtran}


\vfill
\end{document}